\documentclass[runningheads]{llncs}
\usepackage[T1]{fontenc}
\usepackage{graphicx}
\usepackage{subcaption}
\usepackage{caption}
\begin{document}
\title{Aligning Knowledge Graphs Provided by Humans and Generated from Neural Networks in Specific Tasks}
\author{Tangrui Li\inst{1}\orcidID{0009-0005-3471-038X} \and Jun Zhou\inst{2}}
\authorrunning{Li. et al.}
\institute{Temple University, Philadelphia PA 19122, USA \email{tuo90515@temple.edu} \and
Temple University, Philadelphia PA 19122, USA
\email{tun40202@temple.edu} }
\maketitle              % typeset the header of the contribution
\begin{abstract}
This paper develops an innovative method that enables neural networks to generate and utilize knowledge graphs, which describe their concept-level knowledge and optimize network parameters through alignment with human-provided knowledge. This research addresses a gap where traditionally, network-generated knowledge has been limited to applications in downstream symbolic analysis or enhancing network transparency. By integrating a novel autoencoder design with the Vector Symbolic Architecture (VSA), we have introduced auxiliary tasks that support end-to-end training. Our approach eschews traditional dependencies on ontologies or word embedding models, mining concepts from neural networks and directly aligning them with human knowledge. Experiments show that our method consistently captures network-generated concepts that align closely with human knowledge and can even uncover new, useful concepts not previously identified by humans. This plug-and-play strategy not only enhances the interpretability of neural networks but also facilitates the integration of symbolic logical reasoning within these systems.

\end{abstract}

\section{Introduction}

In recent years, more and more studies have focused on specifying the knowledge in neural networks explicitly [1, 2, 3, 7, 8] despite the fact that profound capabilities in managing complex cognitive tasks have already been exhibited [26, 27] for investigating new potentials. Nevertheless, leveraging the knowledge to enhance the training process continues to pose significant challenges. Numerous studies have attempted to derive symbolic knowledge from neural networks, which could potentially be integrated with various knowledge bases to facilitate the training process. For instance, consider a neural network that mistakenly classifies an object as "an apple but not a fruit"; logically, this error should influence the gradient in backpropagation. Existing studies, however, have primarily used this derived knowledge as mere descriptors for one-time network characterization, thus impeding the network’s ability to incorporate logical reasoning progressively.

We propose a new method that utilizes knowledge graphs and Vector Symbolic Architecture (VSA) [28] to implement bidirectional translation between neural network vectors and concept-level knowledge, accompanied by a feedback mechanism that aligns the knowledge of humans and neural networks based on structural similarities. This approach enables the direct application of symbolic AI and logical knowledge bases [19, 20, 21] to support the learning processes. As an example, knowledge graph completion technology [21] can be applied to the knowledge graphs generated by neural networks to enhance their breadth and consistency.

We propose a new autoencoder [29] design, using a tensor representation for the neural network knowledge graph ($KGV_{NN}$) to replace the embedding vector. This not only enables us to acquire knowledge of neural networks through unsupervised learning, but also allows $KGV_{NN}$ to be applied to a variety of downstream tasks without compromising the network’s end-to-end training.

In a nutshell, we: \textbf{a)} propose a method that utilizes knowledge graphs and Vector Symbolic Architecture (VSA) to convert neural network vectors into concept-level knowledge; and \textbf{b)} demonstrate how this knowledge can interact with external knowledge bases with an alignment of concepts generated by neural networks with those provided by humans to enhance both neural network training and interpretability.

The rest of this paper is organized as follows: \textbf{Section 2} describes the related studies and identifies the areas requiring breakthroughs. \textbf{Section 3} first presents background knowledge of VSA then introduces our method in detail. \textbf{Section 4} reports on one concrete experiment and several theoretical experiments designed to assess our method's effectiveness in aligning concepts generated by neural networks with those used by humans, and in applying these alignments for network refinement. Finally, \textbf{Section 5} gives our conclusions and discussions, including our future plan.

\section{Related Works}

Many studies focus on extracting logical knowledge from neural networks though they often fail to exploit it. Instead, this knowledge is merely used to perform downstream logical analysis tasks [1, 3, 7] or to interpret neural networks for transparency [2, 7, 8]. For example, one study [1] extracted fuzzy logic rules and applied them to an inference system to verify their validity, but the performance of these rules was not used for further modifying the neural network. Similarly, in another study [2], the extracted fuzzy logic rules are used only to make a recurrent neural network's functioning interpretable.

Although some studies have tried to use this knowledge for model feedback, they struggled with balancing the \textbf{content} and \textbf{form} of the knowledge. One study [15] developed a group of loss functions to ensure outputs adhered to logical form constraints; however, it failed to capture the concept-level content due to its distributive nature. Conversely, another study [11] used datasets with human-provided explanations [12, 16], in which the content was easily recognizable. Yet, since these datasets often take non-formal logic forms (e.g., natural language, heat maps), they are difficult to process with many existing symbolic reasoning methods. Moreover, assigning a single explanation to any benchmark data item is problematic. Studies [13, 14] have shown that depending on one explanation alone can cause ignorance of certain features and bias the interpretation.

We propose using a knowledge graph as the form of knowledge representation. Firstly, being representable as adjacency matrices, which correlate with tensors in neural networks, it enhances the transparency. Furthermore, by merely constraining the form (not the content), this method remains open to extensive related research [17, 18, 19, 20], enabling seamless integration with previous studies. For instance, it can be combined with the knowledge graph completion technology mentioned above [21].

Except that neural networks can represent knowledge in this form (say $KG_{NN}$), humans can provide knowledge in this form (say $KG_G$) as well. If a match between these two graphs can be established, then the human knowledge can be used as supervision information (with the help of traditional symbolic methods, but not necessary) to optimize the neural network. This does not require the human knowledge to be as specific as in benchmarks. For instance, in image classification, an intuitive and general triplet like "apple, Is-A, fruit" may be provided. Although this type of knowledge might be unsystematic and incomplete, it still offers valuable concepts for analysis. Importantly, utilizing such low-quality knowledge does not preclude the use of more refined one. We restrict only the form (as knowledge graphs), not the content, which allows for the incorporation of other methods for enhancement. While this paper does not explore on it, implementing them is entirely feasible.

To have $KG_{NN}$ is letting neural networks form their own concepts. Although numerous studies have shown that it's feasible to extract these concepts by summarizing internal activities, they often fail to align with human knowledge. For instance, one study [4] identified patterns (vectors) in neural networks as concepts, while others [5, 6, 9] considered neuron activities as concepts; however, all such concepts remain anonymous to us (named as blank nodes). Studies such as research [10], strive to align these anonymous ones with those used by humans by introducing additional training tasks. Consider concepts $s_1$ and $s_2$ representing the numbers 4 and 7, respectively. Then, ideally, $s_1+s_2$ should correspond to the number 11. However, this type of concept matching, especially when restricted to a single task such as integer addition or subtraction, has been criticized to be biased [13, 14]. 

In this case, although we can get $KG_{NN}$, most existing knowledge graph matching methods fail to align it with \( KG_G \) due to blank nodes [24].These methods typically depend on similarity measures derived from ontologies [22] or word embedding models [23]. Methods that can manage blank nodes often involve high computational complexity, as they require matching knowledge graphs' overall structural similarity [25]. In this paper, the knowledge graph generated by neural networks may change frequently and drastically in training, so we cannot afford such methods in the end-to-end structure.

In summary, \textbf{a)} existing studies often fail to use knowledge extracted from neural networks as feedback; \textbf{b)} to address this issue, we utilize knowledge graphs for representation to facilitate matching with human knowledge; \textbf{c)} however, this approach introduces the challenge of blank node matching, which remains unresolved.

\section{Method}
\subsection{Vector Symbolic Architecture Background}

Vector Symbolic Architecture (VSA) uses vector calculations for knowledge representation and query, which are traditionally handled by symbolic systems. In VSA, every atomic symbol is assigned a unique vector, vectors corresponding to different symbols are orthogonal. Through two special operations, bind and bundle, VSA can turn symbolic knowledge into vectors and vice versa.

Let "$*$" represent "bind" in VSA, which is element-wise multiplication, and "$+$" represent "bundle", which is element-wise addition. Given two vectors $A, B$, $A * B$ is a vector perpendicular to $A$ and $B$. $A + B$ is a vector similar to $A$ and $B$ to an extent. In the VSA used in this paper, binding of same symbols can be canceled out, say $[A*B]*A\approx B$. 

For knowledge representation, for example, we can use four symbols, namely $A$, $B$, $Job$, and $Name$, to represent "the job of $A$ is $B$" in a vector, say $K=[A*Name] + [B*Job]$. VSA also supports knowledge query, if we query the $Name$, we can use $K * Name = [[A * Name] + [B * Job]] * Name = [A * Name * Name] + [B * Job * Name] \approx A + [B * Job * Name]$ to get the answer. According to properties of bind and bundle, $K * Name$ is only similar to $A$ (considering atomic symbols).

To support the above properties, VSA needs: \textbf{a)} vectors of different symbols are perpendicular; \textbf{b)} The values in the vectors must only be real numbers, integers or bipolar (-1 and 1) considering different types of VSAs. The VSAs used in this paper are bipolar VSAs.

\subsection{Translation Between $KGV_{NN}$ and $KG_{NN}$}

As a variation of the autoencoder architecture, we replace the embedding vector with $KGV_{NN}$, which is a four-dimensional tensor with values between 0 and 1. The first dimension indicates the batch size, the second dimension indicates the relation (in triplets), and the third and fourth dimensions represent the head and tail entities, respectively. For instance, $KGV_{NN}[a, b, c, d] \geq 0.5$ means that for the $a$-th data item in this batch, it shows that the entities with index $c$ and $d$ are in the relation with index $b$ in $KG_{NN}$. Note that the entities and relations here are anonymous concepts only used by neural networks. 

\subsection{Align $KG_{NN}$ and $KG_G$}

As denoted in [Fig. 1 (left)], $KGV_{NN}$ ($KG_{NN}$) serves specific tasks, which might not be a purpose for $KG_G$. Even though there is overlap, these two knowledge graphs may not match perfectly.

We implemented two bipolar VSAs: $VSA_{NN}$ for $KG_{NN}$ and $VSA_G$ for $KG_G$. Thus triplets in knowledge graphs can be represented as vectors, as discussed in \textbf{Section 3.1}. Given six symbols \( H' \), \( R' \), \( T' \) (keywords for head, relation and tail) and \( A \), \( B \), \( C \) (fillers for head, relation and tail). The triplet \( (A, B, C) \) can be represented as a vector \( K = [H'*A] + [R'*B] + [T'*C] \). This establishes a one-to-one mapping between triplets and vectors, changing our problem to a vector alignment problem between \( KG_{NN} \) and \( KG_G \), and eschews issues the traditional knowledge graph matching methods.

Considering they contain different but overlapping triplets, we chose to use the bipartite matching algorithm [31] to find and enhance similar vectors. Specifically, we employed the loss function \( L_K = \frac{1}{M} \sum (1 - \cos(K_{NN}[i], K_G[j])) \), where \( i \) and \( j \) are indices of paired vectors, and \( M \) is the number of matches. \( K_{NN}[i] = [H' * A] + [R' * B] + [T' * C], K_G[j] = [H' * X] + [R' * Y] + [T' * Z] \) for \( A, B, C \) are symbols in \( VSA_{NN} \) and \( X, Y, Z \) are symbols in \( VSA_G \). Optimization of \( L_K \) occurs in the training of vectors in \( VSA_{NN} \). Subsequently, the focus shifts to aligning vectors of concepts in \( VSA_{NN} \) and \( VSA_G \). As a result, by comparing vectors, we can align concepts with different names.

Additionally, to maintain the validity of \( VSA_{NN} \), we added two regulators: \( L_{R1} = \frac{1}{n_{NN}} \sum | \cos(v[i], v[j]) | \) for \( i \neq j \), \( v[k] \) is the vector of the \( k \)-th symbol in \( VSA_{NN} \), and $n_{NN}$ is the number of symbols in $VSA_{NN}$. \( L_{R1} \) is to make sure all atomic symbols in $VSA_{NN}$ have independent vectors. Another regulator \( L_{R2} = \frac{1}{n_{NN}} \sum \min((v[i]-1)^2, (v[i]+1)^2) \), which penalizes values deviating from -1 or 1. To sum up, \( L_{VSA} = L_K + L_{R1} + L_{R2} \).

Once we have identified the matching concepts in \( KG_{NN} \) and \( KG_G \), $KG_{NN}$ can be trained using the knowledge in $KG_G$ as the label. However, given the potential differences between $KG_{NN}$ and $KG_G$, we only corrected the components in \( KG_{NN} \) that directly conflict with $KG_G$, and the others remained unchanged. This can be treated as a regression task with \( L_R \) as the loss. Since \( KGV_{NN} \) is also utilized for downstream tasks (decoder or any specific tasks), in which the loss is summarized as \(L_T\). Ultimately, we resulted in the loss function \( L = L_{VSA} + L_R + L_T \) for optimizing both the neural network and \( VSA_{NN} \).

In summary, \textbf{a)} we used VSA to convert the knowledge graph matching problem to a vector matching problem, and \textbf{b)} converted it to the concept alignment (what we want) by training \( VSA_{NN} \). We added two regulators to ensure the validity of \( VSA_{NN} \), and trained the neural network with \( KG_G \) as the label.

\begin{figure}[htp]
\centering

\begin{subfigure}[c]{.45\textwidth}
  \centering
  \includegraphics[width=\linewidth]{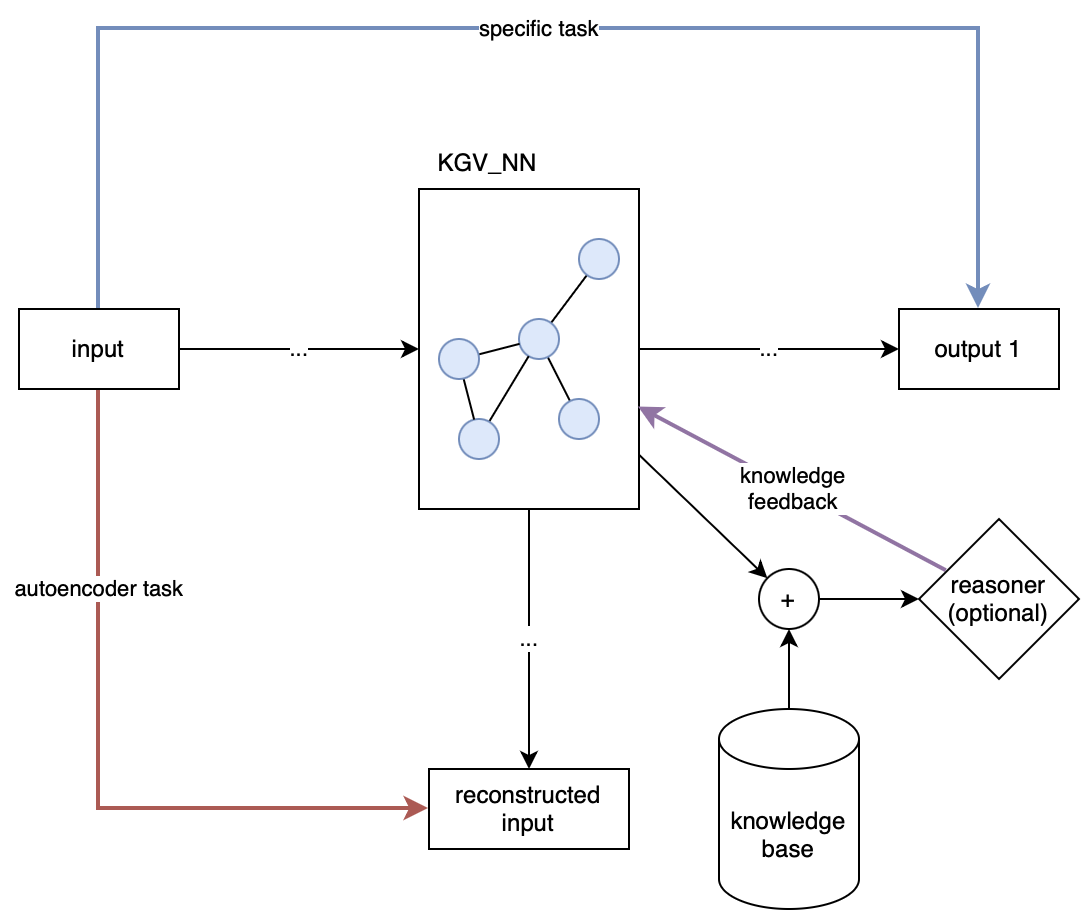}
  \label{fig:image1}
\end{subfigure}\hfill
\begin{subfigure}[c]{.55\textwidth}
  \centering
  \includegraphics[width=\linewidth]{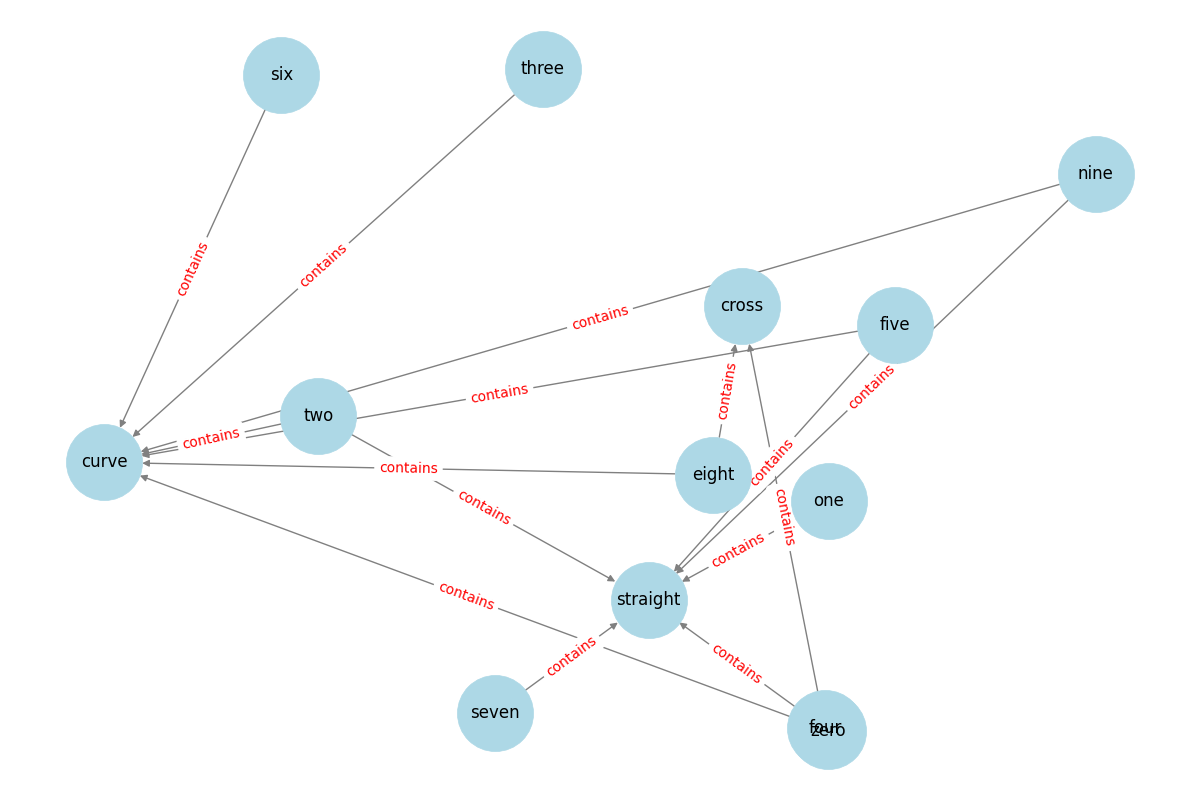}
  \label{fig:image2}
\end{subfigure}
\caption{(left) The structure of the method, in which \( KGV_{NN} \) can be used for downstream tasks, like autoencoder or specific tasks; (right) A sample knowledge graph for MNIST.)}
\label{fig:images}
\end{figure}

\section{Experiments}

We conducted two experiments on MNIST [30] with different \( KG_G \)'s. \( KG_{G1} \) represents the morphological characteristics of handwritten numbers. However, we realized that different people can give different entities, relations, and knowledge. Therefore, in Experiment 2, we tested our method with $KG_G2$ varying in four aspects: \textbf{1)} The number of entities; \textbf{2)} The number of relations; \textbf{3)} The number of triplets; \textbf{4)} Whether the triplets of different relations are evenly given. In addition, we also test our method considering different numbers of concepts in \( KG_{NN} \) given \( KG_G \). Experiment 2 provided guidance for designing neural networks and $KG_G$.

\subsection{Experiment 1}

We designed 15 pieces of knowledge for \( KG_G \) as shown in [Fig. 1 (right)], using 13 entities, which are $zero$, $one$, $two$, $three$, $four$, $five$, $six$, $seven$, $eight$, $nine$, $straight$, $cross$, $curve$, and a relation $contains$. The corresponding bipolar \( VSA_G \) also contains keywords (\( H', R', T' \)). For the bipolar \( VSA_{NN} \), except for keywords (the same as in \( VSA_G \)), we introduced 28 anonymous entities and 1 anonymous relation for it. The following are the results in 10 epochs.

We examined the cosine similarity (consistency) of the matching concepts in \( VSA_{NN} \) and \( VSA_G \), and the two properties of \( VSA_{NN} \) as \( L_{R1} \) and \( L_{R2} \) denoted. In the end, we find all human-provided concepts matched with the average consistency 0.993 (ad shown in [Fig. 2 (e)]). The average similarity between different vectors is 0.001 which showed all atomic symbols in \( VSA_{NN} \) were almost independent. The average bipolar loss (\( L_{R2} \)) was 0.027, indicating in majority the values in vectors were -1 and 1. The test set classification accuracy was 0.972. 

Note that in getting the consistency, the matching between concepts in \( KG_{NN} \) and \( KG_G \) was given by the bipartite matching algorithm with no thresholds. When there are fewer concepts in \( KG_G \), we will find a match for all of them in \( KG_{NN} \), and vice versa.

\begin{figure}[htp]
  \centering
  % Left block, center aligned
  \begin{minipage}[c]{0.45\textwidth}
    \centering
    % Image a
    \begin{subfigure}[b]{0.48\textwidth}
      \centering
      \includegraphics[width=\linewidth]{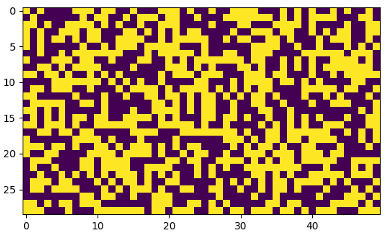}
      \caption*{\centering (a)} % Explicitly centered label for image a
    \end{subfigure}
    \hfill
    % Image b
    \begin{subfigure}[b]{0.48\textwidth}
      \centering
      \includegraphics[width=\linewidth]{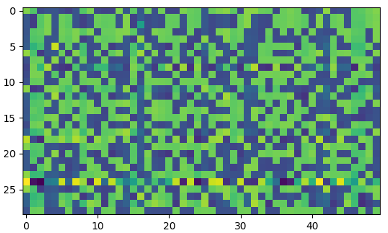}
      \caption*{\centering (b)} % Explicitly centered label for image b
    \end{subfigure}

    % Image c
    \begin{subfigure}[b]{0.48\textwidth}
      \centering
      \includegraphics[width=\linewidth]{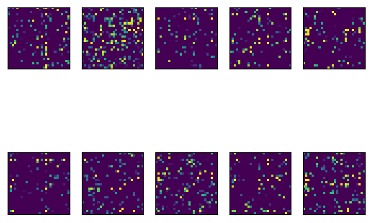}
      \caption*{\centering (c)} % Explicitly centered label for image c
    \end{subfigure}
    \hfill
    % Image d
    \begin{subfigure}[b]{0.48\textwidth}
      \centering
      \includegraphics[width=\linewidth]{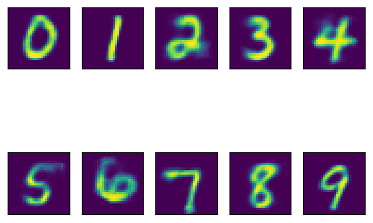}
      \caption*{\centering (d)} % Explicitly centered label for image d
    \end{subfigure}
  \end{minipage}
  % Space between left and right blocks
  \hfill
  % Right block, center aligned
  \begin{minipage}[c]{0.53\textwidth}
    \centering
    % Image e
    \begin{subfigure}[b]{\textwidth}
      \centering
      \includegraphics[width=\linewidth]{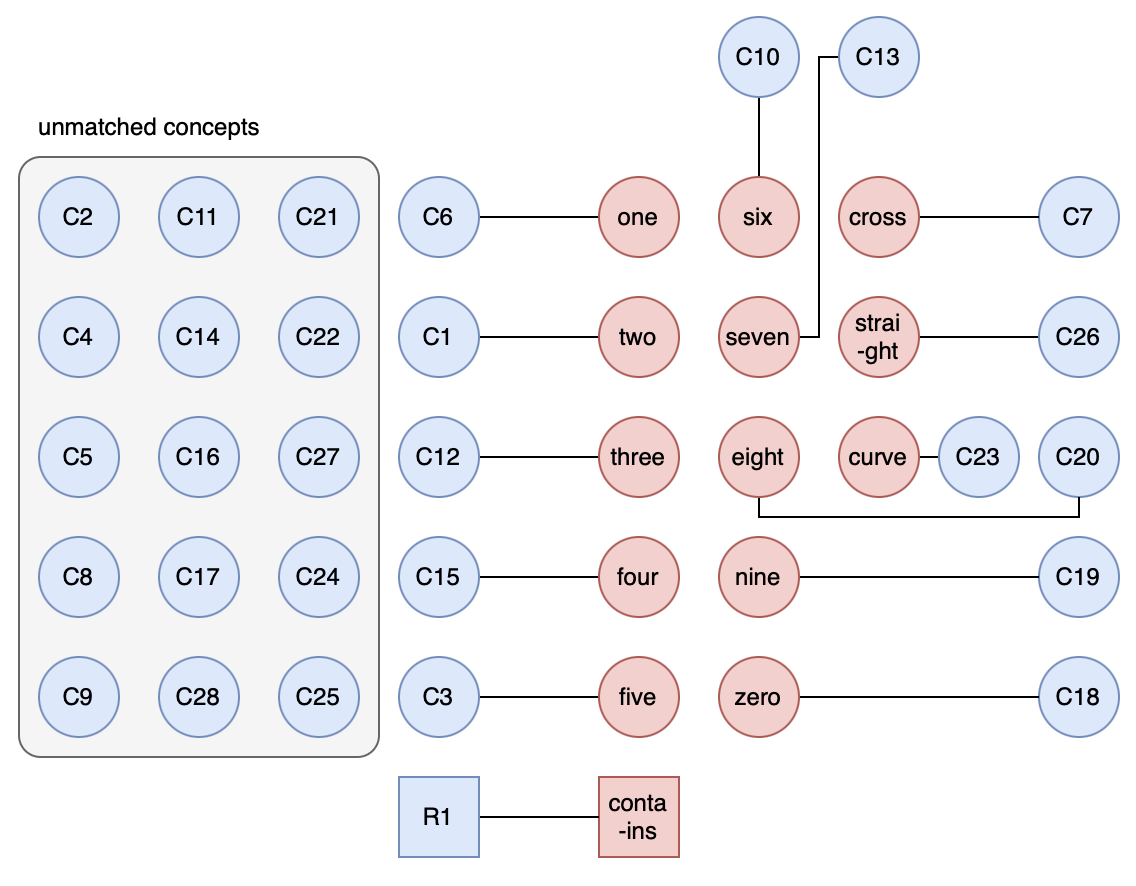}
      \caption*{\centering (e)} % Explicitly centered label for image e
    \end{subfigure}
  \end{minipage}

  \caption{(a) \( VSA_{NN} \) before training, first 50-D; (b) \( VSA_{NN} \) after training, first 50-D; (c) Visualization of the average \( KGV_{NN} \) for all numbers; (d) Visualization of the average \( KGV_{NN} \) for all numbers through the decoder, values $\geq 0.5$; (e) The matching between \( VSA_{NN} \) and \( VSA_G \).}
\end{figure}

\subsection{Experiment 2}

We adjusted \( KG_G \) and \( KG_{NN} \) in five aspects: \textbf{1)} The number of entities in \( KG_G \), with the default value 20; \textbf{2)} The proportion of the relations in entities in \( KG_G \), with the default value 0.2. If there are 20 entities, there will be 4 relations; \textbf{3)} The proportion of triplets in all possible triplets, with the default value 0.2. If there are 20 entities and 4 relations, there will be $20\times 20\times 4=1600$ triplets in total, then by default, \( KG_G \) will have 320 triplets; \textbf{4)} Discussing whether the number of triplets for different relations is uniformly distributed; \textbf{5)} The proportion of the number of entities and relations in \( KG_{NN} \) to \( KG_G \), with the default value 1. If there are 20 entities and 4 relations in \( KG_G \), there will also be 20 entities and 4 relations in \( KG_{NN} \). In Experiment 2, when one aspect is examined, the default values are used for all other aspects. We will also focus on three evaluations used in Experiment 1, say consistency, similarity, and Bipolar loss.

When examining the number of entities (say $\#e$) in \( KG_G \), we found that usually the larger the $\#e$ the better the matching, as shown in [Fig. 3 (a)]. When examining the proportion of relations to entities (say $\%r$), usually the larger the $\%r$ the better the matching, as shown in [Fig. 3 (b)]. When examining the number of triplets (say $\%k$), usually the more abundant the $\%k$, the better the matching, as shown in [Fig. 3 (c)]. When examining whether the triplets are distributed uniformly among relations, we introduced $\alpha\in(0,1)$, and used a normal distribution with variance $0.1 * (1 - \alpha) + 0.01 * \alpha$, mean 0.5 to generate the proportion of triplets covered in each relation. For instance, 0.7 means only 70\% of the triplets in this relation is used. We found that when the knowledge was not concentrated on only a few relations, there was a better matching, as shown in [Fig. 3 (d)]. Finally, when examining the number of concepts in \(KG_{NN}\), we found that usually the more the concepts in \( KG_{NN} \), the better the matching, as shown in [Fig. 3 (e)]. And this aspect has the greatest impact on the consistency. It is also consistent with our conjecture that there are anonymous concepts for down stream tasks in \( KG_{NN} \).

\begin{figure}[htp]
  \centering
  % First row of images
  \begin{subfigure}[b]{0.32\textwidth}
    \centering
    \includegraphics[width=\textwidth]{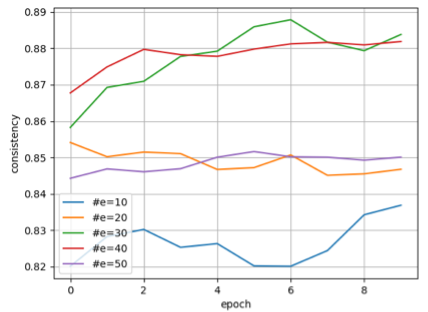}
    \caption*{\centering(a)}
    \label{fig:first}
  \end{subfigure}
  \hfill
  \begin{subfigure}[b]{0.32\textwidth}
    \centering
    \includegraphics[width=\textwidth]{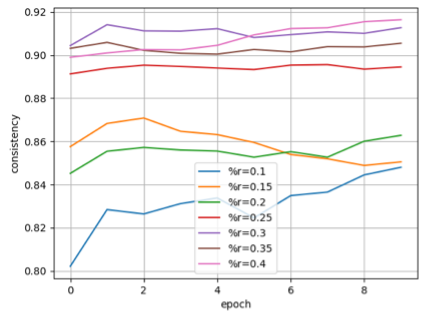}
    \caption*{\centering(b)}
    \label{fig:second}
  \end{subfigure}
  \hfill
  \begin{subfigure}[b]{0.32\textwidth}
    \centering
    \includegraphics[width=\textwidth]{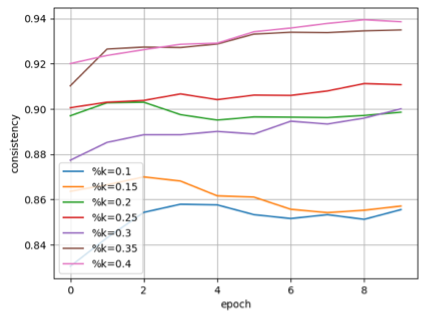}
    \caption*{\centering(c)}
    \label{fig:third}
  \end{subfigure}

  % Second row of images, centered and together
  \centering % This ensures that the group of images is centered
  \begin{subfigure}[b]{0.32\textwidth}
    \centering
    \includegraphics[width=\textwidth]{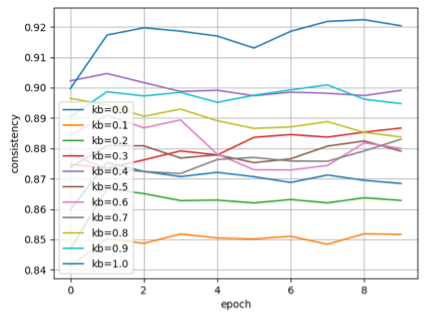}
    \caption*{\centering(d)}
    \label{fig:fourth}
  \end{subfigure}
  % No space command here, images will be adjacent
  \begin{subfigure}[b]{0.32\textwidth}
    \centering
    \includegraphics[width=\textwidth]{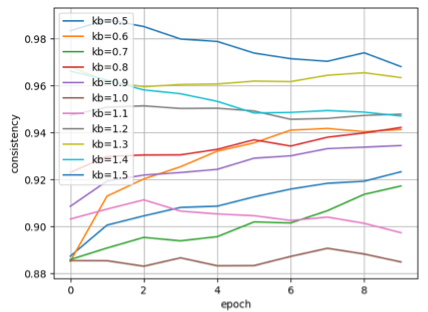}
    \caption*{\centering(e)}
    \label{fig:fifth}
  \end{subfigure}

  \caption{Experiment 2 results. All of them show the average cosine similarity of matched vectors in 10 epochs. (a) Takes the number of entities in \(KG_G\) as the variable. (b) Takes the number of relations in \(KG_G\) as the variable. (c) Takes the number of knowledge (triplets) in \(KG_G\) as the variable. (d) Takes the variance of triplets among relations in \(KG_G\) as the variable. (e) Takes the number of concepts (entities/relations) in \( KG_{NN} \) considering the number of concepts in \( KG_G \) as the variable.}
  \label{fig:images}
\end{figure}

\subsection{Experiment Summary}

Our experiments demonstrated the effectiveness of the alternated autoencoder and concept matching methods through a concrete example. Further tests examined the performance in different neural network parameters and human knowledge specifications. The results showed that in most cases, we got very good matching. This method performed best especially when the knowledge content provided by humans is rich, there were many entities and relationships, the triplets were evenly distributed across relations, and \( VSA_{NN} \) was large.

\section{Conclusion}

The proposed method successfully verified the use of Vector Symbol Architecture (VSA) for matching knowledge graphs, and demonstrated its applicability in neural network concept generation and alignment. Experiments showed that our approach could work in a variety of situations with a high matching consistency. This method not only improves the transparency of neural networks, but also demonstrates the potential of combining neural network models with existing knowledge graph processing technologies. All codes are available at https://github.com/MoonWalker1997/VSANNKGAlignment.

However, despite these enlightening results, the experiments’ complexity remains limited, and the underlying theories require further validation in more complex practical scenarios. In the future, we plan to explore the integration of more symbolic artificial intelligence methods.


\end{document}